\documentclass[10pt,twocolumn,letterpaper]{article}

\usepackage{cvpr}              

\usepackage{graphicx}
\usepackage{amsmath}
\usepackage{amssymb}
\usepackage{booktabs}
\usepackage{multirow}
\usepackage{makecell}
\usepackage{xspace}
\usepackage{colortbl}
\usepackage{tikz}
\usetikzlibrary{arrows.meta,positioning,fit,backgrounds,calc,shapes.geometric,shapes.misc,decorations.pathreplacing,shadows}
\usepackage{bm}
\usepackage{pifont}

\newcommand{\method}{ExecHMR\xspace}

\newcommand{\qdes}{\mathbf{q}^{\mathrm{des}}}
\newcommand{\hq}{\hat{\mathbf{q}}^{\mathrm{des}}}
\newcommand{\cL}{\mathcal{L}}
\newcommand{\R}{\mathbb{R}}

\newcommand{\best}[1]{\textbf{#1}}
\newcommand{\second}[1]{\underline{#1}}

\renewcommand{\paragraph}[1]{\vspace{.3em}\noindent\textbf{#1}\hspace{.4em}}

\usepackage{fix-cm}

\definecolor{cvprblue}{rgb}{0.21,0.49,0.74}
\usepackage[pagebackref,breaklinks,colorlinks,allcolors=cvprblue]{hyperref}

\def\confName{CVPR}
\def\confYear{2027}

\AddToHook{file/1_intro.tex/before}{\colorlet{savedred}{red}\colorlet{red}{black}}
\AddToHook{file/1_intro.tex/after}{\colorlet{red}{savedred}}

\title{\method: Transition-Aware Recovery of Robot-Executable\\Human Motion with Bidirectional Alignment}

\author{
Jamie, Leo Fung, Jack Yu, and Zhao Wang$^{\dagger *}$\\
LimX Dynamics\\
$^{\dagger}$Project Lead \quad $^{*}$Corresponding Author
}

\begin{document}
\maketitle
\newcommand{\numLabelFrames}{29.9M}
\newcommand{\numLabelAcc}{83.0}
\newcommand{\numLabelSeq}{118{,}357}
\newcommand{\numActMaeDeg}{6.25}
\newcommand{\numActRmseSim}{0.165}
\newcommand{\numActRmseRet}{0.213}
\newcommand{\numActNmse}{0.202}
\newcommand{\numActStepZero}{0.1744}
\newcommand{\numActStepFour}{0.1750}
\newcommand{\numActSimFrac}{82.4}
\newcommand{\numSuccGt}{70.4}
\newcommand{\numFeasGt}{100.0}
\newcommand{\numSuccGenmo}{71.3}
\newcommand{\numFeasGenmo}{90.5}
\newcommand{\numSuccBase}{64.1}
\newcommand{\numFeasBase}{83.7}
\newcommand{\numSuccVel}{62.4}
\newcommand{\numFeasVel}{82.4}
\newcommand{\numSuccVelfoots}{64.6}
\newcommand{\numFeasVelfoots}{84.5}
\newcommand{\numSuccOurss}{65.4}
\newcommand{\numFeasOurss}{84.8}
\newcommand{\numSuccOursl}{65.5}
\newcommand{\numFeasOursl}{85.6}

\newcommand{\corrpampjpe}{$-0.02$}
\newcommand{\corrmpjpe}{$+0.05$}
\newcommand{\corrwmpjpe}{$+0.32$}
\newcommand{\corrrte}{$-0.04$}
\newcommand{\corrjitter}{$+0.22$}
\newcommand{\corrfootsliding}{$+0.22$}
\newcommand{\corrPairs}{490}

\newcommand{\sAllGt}{77.7}
\newcommand{\sFeasGt}{100.0}
\newcommand{\sSubGt}{78.2}
\newcommand{\sSubFeasGt}{100.0}
\newcommand{\sERGt}{76.6}
\newcommand{\sERFeasGt}{100.0}
\newcommand{\sAllWham}{73.8}
\newcommand{\sFeasWham}{82.6}
\newcommand{\sSubWham}{75.5}
\newcommand{\sSubFeasWham}{81.8}
\newcommand{\sERWham}{69.7}
\newcommand{\sERFeasWham}{80.7}
\newcommand{\sAllGvhmr}{79.0}
\newcommand{\sFeasGvhmr}{89.2}
\newcommand{\sSubGvhmr}{80.6}
\newcommand{\sSubFeasGvhmr}{88.3}
\newcommand{\sERGvhmr}{75.3}
\newcommand{\sERFeasGvhmr}{87.6}
\newcommand{\sAllTram}{75.6}
\newcommand{\sFeasTram}{87.8}
\newcommand{\sSubTram}{76.6}
\newcommand{\sSubFeasTram}{86.6}
\newcommand{\sERTram}{72.7}
\newcommand{\sERFeasTram}{88.4}
\newcommand{\sAllDuomo}{76.1}
\newcommand{\sFeasDuomo}{86.6}
\newcommand{\sSubDuomo}{77.2}
\newcommand{\sSubFeasDuomo}{85.2}
\newcommand{\sERDuomo}{71.4}
\newcommand{\sERFeasDuomo}{85.4}
\newcommand{\sAllGem}{79.7}
\newcommand{\sFeasGem}{89.2}
\newcommand{\sSubGem}{80.4}
\newcommand{\sSubFeasGem}{87.6}
\newcommand{\sERGem}{76.0}
\newcommand{\sERFeasGem}{88.0}
\newcommand{\sAllFlowhmr}{76.3}
\newcommand{\sFeasFlowhmr}{86.0}
\newcommand{\sSubFlowhmr}{77.2}
\newcommand{\sSubFeasFlowhmr}{84.2}
\newcommand{\sERFlowhmr}{73.0}
\newcommand{\sERFeasFlowhmr}{84.1}
\newcommand{\sAllOurs}{77.0}
\newcommand{\sFeasOurs}{87.8}
\newcommand{\sSubOurs}{78.2}
\newcommand{\sSubFeasOurs}{86.6}
\newcommand{\sEROurs}{73.0}
\newcommand{\sERFeasOurs}{85.4}
\newcommand{\sAllOursAct}{86.0}
\newcommand{\sFeasOursAct}{95.6}
\newcommand{\sSubOursAct}{89.2}
\newcommand{\sSubFeasOursAct}{96.2}
\newcommand{\sEROursAct}{81.2}
\newcommand{\sERFeasOursAct}{93.6}
\newcommand{\nSegSub}{372}
\newcommand{\nSegFlow}{443}
\newcommand{\nSegRichFlow}{71}
\newcommand{\nVidRichFlow}{33}
\newcommand{\sFlowGt}{77.7}
\newcommand{\sFlowFeasGt}{100.0}
\newcommand{\sRichFlowGt}{74.6}
\newcommand{\sFlowWham}{73.8}
\newcommand{\sFlowFeasWham}{82.6}
\newcommand{\sRichFlowWham}{64.8}
\newcommand{\sFlowGvhmr}{79.0}
\newcommand{\sFlowFeasGvhmr}{89.2}
\newcommand{\sRichFlowGvhmr}{70.4}
\newcommand{\sFlowTram}{75.6}
\newcommand{\sFlowFeasTram}{87.8}
\newcommand{\sRichFlowTram}{70.4}
\newcommand{\sFlowDuomo}{76.1}
\newcommand{\sFlowFeasDuomo}{86.6}
\newcommand{\sRichFlowDuomo}{70.4}
\newcommand{\sFlowGem}{79.7}
\newcommand{\sFlowFeasGem}{89.2}
\newcommand{\sRichFlowGem}{76.1}
\newcommand{\sFlowFlowhmr}{76.3}
\newcommand{\sFlowFeasFlowhmr}{86.0}
\newcommand{\sRichFlowFlowhmr}{71.8}
\newcommand{\sFlowOurs}{77.0}
\newcommand{\sFlowFeasOurs}{87.8}
\newcommand{\sRichFlowOurs}{70.4}
\newcommand{\sFlowOursAct}{86.0}
\newcommand{\sFlowFeasOursAct}{95.6}
\newcommand{\sRichFlowOursAct}{69.0}
\newcommand{\nSegER}{304}
\newcommand{\duomoPartial}[2]{#2}
\newcommand{\nSegAll}{443}
\newcommand{\nVidBench}{112}
\newcommand{\sGainOursAct}{9.0}
\newcommand{\nSegFeas}{344}
\newcommand{\pwOursActOurs}{41}
\newcommand{\plOursActOurs}{1}
\newcommand{\ppOursActOurs}{<10^{-4}}
\newcommand{\pvOursActOurs}{significantly better}
\newcommand{\pwOursActWham}{70}
\newcommand{\plOursActWham}{16}
\newcommand{\ppOursActWham}{<10^{-4}}
\newcommand{\pvOursActWham}{significantly better}
\newcommand{\pwOursActGvhmr}{48}
\newcommand{\plOursActGvhmr}{17}
\newcommand{\ppOursActGvhmr}{<10^{-3}}
\newcommand{\pvOursActGvhmr}{significantly better}
\newcommand{\pwOursActTram}{58}
\newcommand{\plOursActTram}{12}
\newcommand{\ppOursActTram}{<10^{-4}}
\newcommand{\pvOursActTram}{significantly better}
\newcommand{\pwOursActDuomo}{54}
\newcommand{\plOursActDuomo}{10}
\newcommand{\ppOursActDuomo}{<10^{-4}}
\newcommand{\pvOursActDuomo}{significantly better}
\newcommand{\pwOursActGem}{45}
\newcommand{\plOursActGem}{17}
\newcommand{\ppOursActGem}{<10^{-3}}
\newcommand{\pvOursActGem}{significantly better}
\newcommand{\pwOursActFlowhmr}{61}
\newcommand{\plOursActFlowhmr}{18}
\newcommand{\ppOursActFlowhmr}{<10^{-4}}
\newcommand{\pvOursActFlowhmr}{significantly better}
\newcommand{\pwOursWham}{44}
\newcommand{\plOursWham}{30}
\newcommand{\ppOursWham}{=0.13}
\newcommand{\pvOursWham}{not significant}
\newcommand{\pwOursGvhmr}{20}
\newcommand{\plOursGvhmr}{29}
\newcommand{\ppOursGvhmr}{=0.25}
\newcommand{\pvOursGvhmr}{not significant}
\newcommand{\pwOursTram}{36}
\newcommand{\plOursTram}{30}
\newcommand{\ppOursTram}{=0.54}
\newcommand{\pvOursTram}{not significant}
\newcommand{\pwOursDuomo}{32}
\newcommand{\plOursDuomo}{28}
\newcommand{\ppOursDuomo}{=0.70}
\newcommand{\pvOursDuomo}{not significant}
\newcommand{\pwOursGem}{19}
\newcommand{\plOursGem}{31}
\newcommand{\ppOursGem}{=0.12}
\newcommand{\pvOursGem}{not significant}
\newcommand{\pwOursFlowhmr}{30}
\newcommand{\plOursFlowhmr}{27}
\newcommand{\ppOursFlowhmr}{=0.79}
\newcommand{\pvOursFlowhmr}{not significant}
\newcommand{\sActBlend}{5.6}
\newcommand{\sActBlendTwenty}{52.4}
\newcommand{\sActInit}{77.4}
\newcommand{\sActOpenloop}{0.0}
\newcommand{\numActStep}{50k}
\newcommand{\numActLogStep}{50k}
\renewcommand{\numActMaeDeg}{5.49}
\renewcommand{\numActRmseSim}{0.147}
\renewcommand{\numActRmseRet}{0.171}
\renewcommand{\numActNmse}{0.160}
\renewcommand{\numActStepZero}{0.1518}
\renewcommand{\numActStepFour}{0.1526}
\renewcommand{\numActSimFrac}{82.3}

\newcommand{\abBase}{76.7}
\newcommand{\abFeasBase}{86.9}
\newcommand{\abVel}{80.6}
\newcommand{\abFeasVel}{91.0}
\newcommand{\abFoot}{80.8}
\newcommand{\abFeasFoot}{91.3}
\newcommand{\abAct}{80.4}
\newcommand{\abFeasAct}{90.1}
\newcommand{\abActRobot}{87.4}
\newcommand{\abFeasActRobot}{95.6}
\newcommand{\abJitBase}{25.3}
\newcommand{\abJitVel}{12.1}
\newcommand{\abJitFoot}{14.8}
\newcommand{\abJitAct}{14.7}
\newcommand{\abwVelBase}{26}
\newcommand{\ablVelBase}{9}
\newcommand{\abpVelBase}{=0.006}
\newcommand{\abwFootVel}{14}
\newcommand{\ablFootVel}{13}
\newcommand{\abpFootVel}{=1.00}
\newcommand{\abwActFoot}{14}
\newcommand{\ablActFoot}{16}
\newcommand{\abpActFoot}{=0.86}
\newcommand{\abwRobotAct}{38}
\newcommand{\ablRobotAct}{7}
\newcommand{\abpRobotAct}{<10^{-4}}
\newcommand{\exRawHuman}{67.9}
\newcommand{\exGfHuman}{77.0}
\newcommand{\exRawRobot}{77.9}
\newcommand{\exGfRobot}{86.0}

\newcommand{\labN}{40}
\newcommand{\labDesHip}{6.9}
\newcommand{\labActHip}{6.4}
\newcommand{\labDesKnee}{8.9}
\newcommand{\labActKnee}{6.6}
\newcommand{\labDesWaist}{7.0}
\newcommand{\labActWaist}{4.1}
\newcommand{\labDesArm}{6.5}
\newcommand{\labActArm}{3.7}
\newcommand{\labDesHead}{3.0}
\newcommand{\labActHead}{2.1}

\newcommand{\intensN}{443}
\newcommand{\intensJointGTSlow}{78.4}
\newcommand{\intensJointGTMid}{81.6}
\newcommand{\intensJointGTFast}{73.0}
\newcommand{\intensJointGVHMRSlow}{86.5}
\newcommand{\intensJointGVHMRMid}{86.4}
\newcommand{\intensJointGVHMRFast}{64.2}
\newcommand{\intensJointGENMOSlow}{85.1}
\newcommand{\intensJointGENMOMid}{87.8}
\newcommand{\intensJointGENMOFast}{66.2}
\newcommand{\intensJointOursHumanModeSlow}{80.4}
\newcommand{\intensJointOursHumanModeMid}{83.0}
\newcommand{\intensJointOursHumanModeFast}{67.6}
\newcommand{\intensJointOursRobotModeSlow}{83.1}
\newcommand{\intensJointOursRobotModeMid}{89.8}
\newcommand{\intensJointOursRobotModeFast}{85.1}
\newcommand{\intensJointLo}{0.46}
\newcommand{\intensJointHi}{0.62}
\newcommand{\intensJointFastwGem}{34}
\newcommand{\intensJointFastlGem}{6}
\newcommand{\intensJointFastpGem}{<10^{-4}}
\newcommand{\intensJointFastwHuman}{27}
\newcommand{\intensJointFastlHuman}{1}
\newcommand{\intensJointFastpHuman}{<10^{-4}}
\newcommand{\intensRootGTSlow}{77.0}
\newcommand{\intensRootGTMid}{81.0}
\newcommand{\intensRootGTFast}{75.0}
\newcommand{\intensRootGVHMRSlow}{78.4}
\newcommand{\intensRootGVHMRMid}{87.8}
\newcommand{\intensRootGVHMRFast}{70.9}
\newcommand{\intensRootGENMOSlow}{77.7}
\newcommand{\intensRootGENMOMid}{88.4}
\newcommand{\intensRootGENMOFast}{73.0}
\newcommand{\intensRootOursHumanModeSlow}{74.3}
\newcommand{\intensRootOursHumanModeMid}{85.0}
\newcommand{\intensRootOursHumanModeFast}{71.6}
\newcommand{\intensRootOursRobotModeSlow}{76.4}
\newcommand{\intensRootOursRobotModeMid}{90.5}
\newcommand{\intensRootOursRobotModeFast}{91.2}
\newcommand{\intensRootLo}{0.41}
\newcommand{\intensRootHi}{0.82}
\newcommand{\intensRootFastwGem}{33}
\newcommand{\intensRootFastlGem}{6}
\newcommand{\intensRootFastpGem}{<10^{-4}}
\newcommand{\intensRootFastwHuman}{29}
\newcommand{\intensRootFastlHuman}{0}
\newcommand{\intensRootFastpHuman}{<10^{-4}}
\newcommand{\evStandard}{77.0}
\newcommand{\evFeasStandard}{87.8}
\newcommand{\evWbtSmoothTwo}{79.5}
\newcommand{\evFeasWbtSmoothTwo}{90.1}
\newcommand{\evwWbtSmoothTwo}{18}
\newcommand{\evlWbtSmoothTwo}{7}
\newcommand{\evWbtSmoothFour}{85.8}
\newcommand{\evFeasWbtSmoothFour}{95.1}
\newcommand{\evwWbtSmoothFour}{41}
\newcommand{\evlWbtSmoothFour}{2}
\newcommand{\evActInit}{77.4}
\newcommand{\evFeasActInit}{88.4}
\newcommand{\evwActInit}{8}
\newcommand{\evlActInit}{6}
\newcommand{\evActRef}{86.0}
\newcommand{\evFeasActRef}{95.6}
\newcommand{\evwActRef}{41}
\newcommand{\evlActRef}{1}
\newcommand{\evActRefKOne}{84.9}
\newcommand{\evFeasActRefKOne}{94.2}
\newcommand{\evwActRefKOne}{38}
\newcommand{\evlActRefKOne}{3}
\newcommand{\evActRefSmoothFour}{86.9}
\newcommand{\evFeasActRefSmoothFour}{95.9}
\newcommand{\evwActRefSmoothFour}{48}
\newcommand{\evlActRefSmoothFour}{4}
\newcommand{\evActBlendTwoZero}{52.4}
\newcommand{\evFeasActBlendTwoZero}{61.3}
\newcommand{\evwActBlendTwoZero}{2}
\newcommand{\evlActBlendTwoZero}{111}
\newcommand{\evActBlend}{5.6}
\newcommand{\evFeasActBlend}{7.0}
\newcommand{\evwActBlend}{0}
\newcommand{\evlActBlend}{316}
\newcommand{\evActOpenloop}{0.0}
\newcommand{\evFeasActOpenloop}{0.0}
\newcommand{\evwActOpenloop}{0}
\newcommand{\evlActOpenloop}{341}
\newcommand{\evOtherWham}{83.5}
\newcommand{\evFeasOtherWham}{91.0}
\newcommand{\evwOtherWham}{46}
\newcommand{\evlOtherWham}{3}
\newcommand{\evOtherGvhmr}{88.5}
\newcommand{\evFeasOtherGvhmr}{97.1}
\newcommand{\evwOtherGvhmr}{47}
\newcommand{\evlOtherGvhmr}{5}
\newcommand{\evOtherTram}{82.2}
\newcommand{\evFeasOtherTram}{93.0}
\newcommand{\evwOtherTram}{32}
\newcommand{\evlOtherTram}{3}
\newcommand{\evOtherDuomo}{85.6}
\newcommand{\evFeasOtherDuomo}{94.8}
\newcommand{\evwOtherDuomo}{43}
\newcommand{\evlOtherDuomo}{1}
\newcommand{\evOtherGem}{88.5}
\newcommand{\evFeasOtherGem}{96.8}
\newcommand{\evwOtherGem}{41}
\newcommand{\evlOtherGem}{2}
\newcommand{\evOtherFlowhmr}{87.8}
\newcommand{\evFeasOtherFlowhmr}{97.1}
\newcommand{\evwOtherFlowhmr}{51}
\newcommand{\evlOtherFlowhmr}{0}
\newcommand{\smRescueRobot}{41}
\newcommand{\smRescueSmooth}{41}
\newcommand{\smRescueBoth}{36}
\newcommand{\smwOursActVsGem}{8}
\newcommand{\smlOursActVsGem}{15}
\newcommand{\smpOursActVsGem}{=0.21}
\newcommand{\smwOursActRawVsGem}{8}
\newcommand{\smlOursActRawVsGem}{19}
\newcommand{\smpOursActRawVsGem}{=0.052}
\newcommand{\smwOursActVsGvhmr}{9}
\newcommand{\smlOursActVsGvhmr}{16}
\newcommand{\smpOursActVsGvhmr}{=0.23}
\newcommand{\smwOursActRawVsGvhmr}{8}
\newcommand{\smlOursActRawVsGvhmr}{19}
\newcommand{\smpOursActRawVsGvhmr}{=0.052}
\newcommand{\smStdWham}{73.8}
\newcommand{\smSmWham}{83.5}
\newcommand{\smwWham}{46}
\newcommand{\smlWham}{3}
\newcommand{\smpWham}{<10^{-4}}
\newcommand{\smStdGvhmr}{79.0}
\newcommand{\smSmGvhmr}{88.5}
\newcommand{\smwGvhmr}{47}
\newcommand{\smlGvhmr}{5}
\newcommand{\smpGvhmr}{<10^{-4}}
\newcommand{\smStdTram}{75.6}
\newcommand{\smSmTram}{82.2}
\newcommand{\smwTram}{32}
\newcommand{\smlTram}{3}
\newcommand{\smpTram}{<10^{-4}}
\newcommand{\smStdDuomo}{76.1}
\newcommand{\smSmDuomo}{85.6}
\newcommand{\smwDuomo}{43}
\newcommand{\smlDuomo}{1}
\newcommand{\smpDuomo}{<10^{-4}}
\newcommand{\smStdGem}{79.7}
\newcommand{\smSmGem}{88.5}
\newcommand{\smwGem}{41}
\newcommand{\smlGem}{2}
\newcommand{\smpGem}{<10^{-4}}
\newcommand{\smStdFlowhmr}{76.3}
\newcommand{\smSmFlowhmr}{87.8}
\newcommand{\smwFlowhmr}{51}
\newcommand{\smlFlowhmr}{0}
\newcommand{\smpFlowhmr}{<10^{-4}}
\newcommand{\smStdOurs}{77.0}
\newcommand{\smSmOurs}{85.8}
\newcommand{\smwOurs}{41}
\newcommand{\smlOurs}{2}
\newcommand{\smpOurs}{<10^{-4}}
\newcommand{\smStdOursAct}{86.0}
\newcommand{\smSmOursAct}{86.9}
\newcommand{\smwOursAct}{7}
\newcommand{\smlOursAct}{3}
\newcommand{\smpOursAct}{=0.34}

\newcommand{\ctlMae}{7.42}
\newcommand{\oursMae}{5.49}
\newcommand{\ctlHuman}{79.0}
\newcommand{\ctlRobot}{87.1}
\newcommand{\ctlW}{382.1}
\newcommand{\ctlWA}{143.1}
\newcommand{\ctlRobotW}{10}
\newcommand{\ctlRobotL}{15}
\newcommand{\ctlRobotP}{=0.42}
\newcommand{\ctlHumanW}{16}
\newcommand{\ctlHumanL}{25}
\newcommand{\ctlHumanP}{=0.21}

\newcommand{\abNaMae}{5.49}
\newcommand{\abNaNmse}{0.160}
\newcommand{\abNaW}{323.4}
\newcommand{\abNaWA}{127.5}
\newcommand{\abNaRte}{2.8}
\newcommand{\abNaHuman}{76.1}
\newcommand{\abNaRobot}{85.8}
\newcommand{\abNaTdpwPA}{34.8}
\newcommand{\abNaTdpw}{54.8}
\newcommand{\abNaRichPA}{38.5}
\newcommand{\abNaRich}{66.3}
\newcommand{\abNaEmdbPA}{40.9}
\newcommand{\abNaEmdb}{65.2}
\newcommand{\abNaHumanW}{16}
\newcommand{\abNaHumanL}{12}
\newcommand{\abNaHumanP}{=0.57}
\newcommand{\abNaRobotW}{7}
\newcommand{\abNaRobotL}{6}
\newcommand{\abNaRobotP}{=1.00}
\newcommand{\abNaMaeGain}{0.1}
\newcommand{\abNaWGain}{-2.2}
\newcommand{\abDtMae}{5.61}
\newcommand{\abDtNmse}{0.166}
\newcommand{\abDtW}{330.6}
\newcommand{\abDtWA}{128.7}
\newcommand{\abDtRte}{2.9}
\newcommand{\abDtHuman}{76.3}
\newcommand{\abDtRobot}{85.3}
\newcommand{\abDtTdpwPA}{34.5}
\newcommand{\abDtTdpw}{54.4}
\newcommand{\abDtRichPA}{38.4}
\newcommand{\abDtRich}{66.0}
\newcommand{\abDtEmdbPA}{40.7}
\newcommand{\abDtEmdb}{64.1}
\newcommand{\abDtHumanW}{17}
\newcommand{\abDtHumanL}{14}
\newcommand{\abDtHumanP}{=0.72}
\newcommand{\abDtRobotW}{10}
\newcommand{\abDtRobotL}{7}
\newcommand{\abDtRobotP}{=0.63}
\newcommand{\abDtMaeGain}{2.2}
\newcommand{\abDtWGain}{0.1}

\newcommand{\swNaAMae}{5.49}
\newcommand{\swNaATdpwPA}{34.8}
\newcommand{\swNaATdpw}{54.8}
\newcommand{\swNaARichPA}{38.5}
\newcommand{\swNaARich}{66.3}
\newcommand{\swNaAEmdbPA}{40.9}
\newcommand{\swNaAEmdb}{65.2}
\newcommand{\swNaAWA}{127.5}
\newcommand{\swNaAW}{323.4}
\newcommand{\swNaAHuman}{76.1}
\newcommand{\swNaARobot}{85.8}
\newcommand{\swNaBMae}{XX}
\newcommand{\swNaBTdpwPA}{XX}
\newcommand{\swNaBTdpw}{XX}
\newcommand{\swNaBRichPA}{XX}
\newcommand{\swNaBRich}{XX}
\newcommand{\swNaBEmdbPA}{XX}
\newcommand{\swNaBEmdb}{XX}
\newcommand{\swNaBWA}{XX}
\newcommand{\swNaBW}{XX}
\newcommand{\swNaBHuman}{XX}
\newcommand{\swNaBRobot}{XX}
\newcommand{\swDtAMae}{5.61}
\newcommand{\swDtATdpwPA}{34.5}
\newcommand{\swDtATdpw}{54.4}
\newcommand{\swDtARichPA}{38.4}
\newcommand{\swDtARich}{66.0}
\newcommand{\swDtAEmdbPA}{40.7}
\newcommand{\swDtAEmdb}{64.1}
\newcommand{\swDtAWA}{128.7}
\newcommand{\swDtAW}{330.6}
\newcommand{\swDtAHuman}{76.3}
\newcommand{\swDtARobot}{85.3}
\newcommand{\swDtBMae}{XX}
\newcommand{\swDtBTdpwPA}{XX}
\newcommand{\swDtBTdpw}{XX}
\newcommand{\swDtBRichPA}{XX}
\newcommand{\swDtBRich}{XX}
\newcommand{\swDtBEmdbPA}{XX}
\newcommand{\swDtBEmdb}{XX}
\newcommand{\swDtBWA}{XX}
\newcommand{\swDtBW}{XX}
\newcommand{\swDtBHuman}{XX}
\newcommand{\swDtBRobot}{XX}
\newcommand{\swOursAMae}{5.49}
\newcommand{\swOursATdpwPA}{34.3}
\newcommand{\swOursATdpw}{53.7}
\newcommand{\swOursARichPA}{38.7}
\newcommand{\swOursARich}{65.9}
\newcommand{\swOursAEmdbPA}{40.8}
\newcommand{\swOursAEmdb}{64.7}
\newcommand{\swOursAWA}{128.9}
\newcommand{\swOursAW}{330.4}
\newcommand{\swOursAHuman}{77.0}
\newcommand{\swOursARobot}{86.0}
\newcommand{\swOursBMae}{XX}
\newcommand{\swOursBTdpwPA}{XX}
\newcommand{\swOursBTdpw}{XX}
\newcommand{\swOursBRichPA}{XX}
\newcommand{\swOursBRich}{XX}
\newcommand{\swOursBEmdbPA}{XX}
\newcommand{\swOursBEmdb}{XX}
\newcommand{\swOursBWA}{XX}
\newcommand{\swOursBW}{XX}
\newcommand{\swOursBHuman}{XX}
\newcommand{\swOursBRobot}{XX}
\newcommand{\swWOneMae}{XX}
\newcommand{\swWOneTdpwPA}{XX}
\newcommand{\swWOneTdpw}{XX}
\newcommand{\swWOneRichPA}{XX}
\newcommand{\swWOneRich}{XX}
\newcommand{\swWOneEmdbPA}{XX}
\newcommand{\swWOneEmdb}{XX}
\newcommand{\swWOneWA}{XX}
\newcommand{\swWOneW}{XX}
\newcommand{\swWOneHuman}{XX}
\newcommand{\swWOneRobot}{XX}
\newcommand{\swWTwoMae}{XX}
\newcommand{\swWTwoTdpwPA}{XX}
\newcommand{\swWTwoTdpw}{XX}
\newcommand{\swWTwoRichPA}{XX}
\newcommand{\swWTwoRich}{XX}
\newcommand{\swWTwoEmdbPA}{XX}
\newcommand{\swWTwoEmdb}{XX}
\newcommand{\swWTwoWA}{XX}
\newcommand{\swWTwoW}{XX}
\newcommand{\swWTwoHuman}{XX}
\newcommand{\swWTwoRobot}{XX}
\newcommand{\swStrTwoMae}{XX}
\newcommand{\swStrTwoTdpwPA}{XX}
\newcommand{\swStrTwoTdpw}{XX}
\newcommand{\swStrTwoRichPA}{XX}
\newcommand{\swStrTwoRich}{XX}
\newcommand{\swStrTwoEmdbPA}{XX}
\newcommand{\swStrTwoEmdb}{XX}
\newcommand{\swStrTwoWA}{XX}
\newcommand{\swStrTwoW}{XX}
\newcommand{\swStrTwoHuman}{XX}
\newcommand{\swStrTwoRobot}{XX}
\newcommand{\swSpreadMae}{XX}
\newcommand{\swSpreadTdpwPA}{XX}
\newcommand{\swSpreadTdpw}{XX}
\newcommand{\swSpreadRichPA}{XX}
\newcommand{\swSpreadRich}{XX}
\newcommand{\swSpreadEmdbPA}{XX}
\newcommand{\swSpreadEmdb}{XX}
\newcommand{\swSpreadWA}{XX}
\newcommand{\swSpreadW}{XX}
\newcommand{\swSpreadHuman}{XX}
\newcommand{\swSpreadRobot}{XX}

\begin{abstract}
Human motion recovery (HMR) from monocular video enables humanoid robot imitation, yet existing methods prioritize pose accuracy without explicitly modeling the transition dynamics required for reliable robot execution.
Recent post-hoc fine-tuning with simulation-based tracking feedback treats recovery fidelity and executability as separate objectives, although human motion and robot action describe the same underlying aligned movement.
We introduce \method, which jointly models human motion recovery and robot action prediction in a single training stage through inter-frame transition reasoning.
\method uses a shared temporal Transformer with human reconstruction and temporal action branches, learning inter-frame transition dynamics for reliable execution from controller commands recorded in closed-loop simulation.
To unify the two branches around this shared movement, bidirectional kinematic alignment matches label-relative residuals in a common kinematic space and alternates teacher roles, enabling mutual guidance while preserving single-pass inference.
On test videos from 3DPW, EMDB and RICH, \method achieves 86.0\% execution success, compared with 79.7\% for GENMO and 76.3\% for FlowHMR, leading all six baselines with the largest gains on fast motions.
It also achieves the lowest camera-space errors on 3DPW and EMDB-1 among compared methods, and we demonstrate deployment on a physical humanoid.
\end{abstract}
\section{Introduction}
\label{sec:intro}

\begin{figure}[t]
\centering
\input{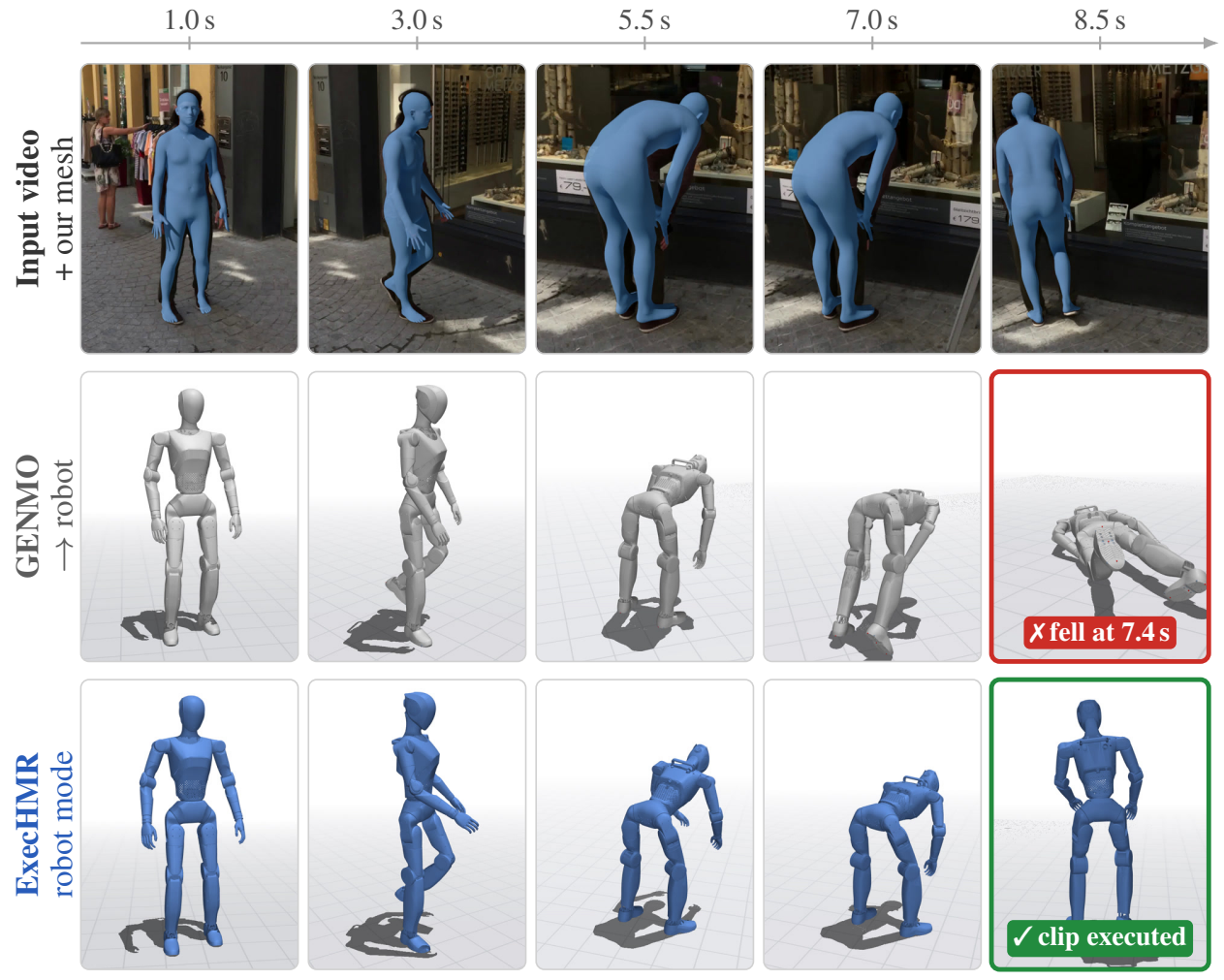}
\vspace{-1.6em}
\caption{
\textbf{Accurate poses do not guarantee executable transitions.}
On this clip, the motion recovered by GENMO~\cite{li2025genmo} causes a fall, while \method completes the sequence using action references learned alongside human motion through transition-aware learning and bidirectional kinematic alignment.
}
\label{fig:teaser}
\vspace{-1.0em}
\end{figure}

Human motion recovery (HMR) from monocular video, a core problem of computer vision, has become a gateway for \textcolor{red}{humanoid} robots to acquire whole-body skills and perform diverse human-like behaviors. 
Motion recovered from videos of almost any human activity can be retargeted to robot and tracked by whole-body tracking (WBT)~\cite{he2024h2o,he2024omnih2o,fu2024humanplus,cheng2024exbody,he2025asap,liao2025beyondmimic,ze2025twist}, turning any video of a human motion \textcolor{red}{trajectory} into a robot demonstration \textcolor{red}{action}~\cite{mao2024uh1,allshire2025videomimic}. 
HMR now reconstructs human body with remarkable accuracy~\cite{shin2024wham,wang2024tram,shen2024gvhmr,li2025genmo,wang2026duomo}, yet humanoid execution requires more than accurate poses in individual \textcolor{red}{frames}: robot must move from one pose to the next while maintaining balance and satisfying \textcolor{red}{dynamic} constraints.
The underlying \textcolor{red}{dynamics} of these \textcolor{red}{inter-frame transitions} are not explicitly modeled by these HMR methods, so visually accurate motions can still be difficult for robots to \textcolor{red}{track and execute} reliably (\cref{fig:teaser}).

Recent methods seek to improve \textcolor{red}{executability} by \textcolor{red}{post-hoc} fine-tuning pre-trained HMR model with simulation-based tracking rewards or preferences~\cite{wang2026flowhmr,yue2025rlpf,ling2026gentrack,he2026vith}.
This \textcolor{red}{subsequent adaptation} may reconcile the objectives of motion fidelity pre-training with the requirements of \textcolor{red}{embodied execution}, potentially introducing a trade-off between \textcolor{red}{fidelity and trackability}.
More fundamentally, human motion and humanoid robot \textcolor{red}{actions} jointly describe comprehensive trajectories that should be faithful to the human demonstration and executable by the robot, yet \textcolor{red}{separate} fine-tuning may trade fidelity for executability without explicitly \textcolor{red}{aligning these representations}.
This motivates jointly \textcolor{red}{modeling} human motion recovery and humanoid robot action prediction within a single model and training stage, with mutual guidance between the two \textcolor{red}{interactive} tasks to capture \textcolor{red}{transition dynamics} while preserving recovery accuracy.

To this end, we introduce ExecHMR, which jointly optimizes recovery fidelity and robot executability in a single training stage by modeling human motion recovery and robot action prediction as coupled tasks through inter-frame transition reasoning.
Through the \textcolor{red}{interaction} between human and robot trajectories, executed humanoid actions provide supervision for learning transition dynamics that support reliable execution, while accurate human motion reconstruction, in turn, guides precise robot action prediction.
We support this \textcolor{red}{interaction} through a shared temporal backbone with task-specific prediction branches, jointly optimized under human motion and executed action supervision.
To reason about trajectory dynamics across these two representations of the same \textcolor{red}{aligned} underlying \textcolor{red}{movement trajectory}, we couple their predictions through \textcolor{red}{mutual constraints} in a shared \textcolor{red}{kinematic} space.

Specifically, we generate humanoid action supervision offline by retargeting the training motions to the humanoid and executing them with its deployed WBT~\cite{fu2026cosa05} in closed-loop simulation, recording continuous joint set-points per transition.
\method then uses the regression Transformer~\cite{li2025genmo,shen2024gvhmr} as a shared temporal backbone, from which two task-specific branches with separate residual adapters predict human motion and robot actions.
The action branch incorporates additional temporal attention to capture the controller's dependence on future reference motion, while supervision from both branches jointly updates the backbone.
To unify both branches under the shared goal of faithfully reconstructing and reliably executing the same underlying motion trajectory, \method leverages \textcolor{red}{bidirectional kinematic alignment} to map predictions and labels through differentiable human and robot kinematics into common body frames and align the resulting prediction residuals.
Two branches alternate as teachers by providing detached residual targets for each other, enabling \textcolor{red}{mutual supervision} while retaining their respective reconstruction and action losses.
At inference, a single forward supports motion recovery and humanoid imitation without motion retarget.

On \nVidBench\ videos from 3DPW, EMDB and RICH, \method achieves \sAllOursAct\% execution success in robot mode, outperforming all six baselines without test-time smoothing (GENMO: \sAllGem\%; FlowHMR: \sAllFlowhmr\%), with the largest gains on fast motions (\cref{tab:robot}).
It also achieves the lowest camera-space errors on 3DPW and EMDB-1 and the lowest EMDB-2 WA-MPJPE (\cref{tab:cam,tab:world}), while reducing training-stream set-point error by 26\% and 3DPW MPJPE by 2\% relative to ablation control, with comparable execution success (\cref{sec:exp_ablation}).
Our contributions are:
\begin{itemize}
\item We propose \method for jointly modeling human motion recovery and robot action prediction through inter-frame transition reasoning and executed-action supervision, coupling recovery fidelity with robot executability in a single training stage.
\item We introduce bidirectional kinematic alignment that couples human and robot predictions through label-relative residuals in a shared kinematic space, enabling mutual supervision while retaining the task-specific losses.
\item We demonstrate \sAllOursAct\% robot execution success without test-time smoothing, the lowest camera-space errors on 3DPW and EMDB-1 among compared methods, and deployment on a physical humanoid.
\end{itemize}
\section{Related Work}
\label{sec:related}

\begin{figure*}[t]
\centering
\includegraphics[width=\textwidth]{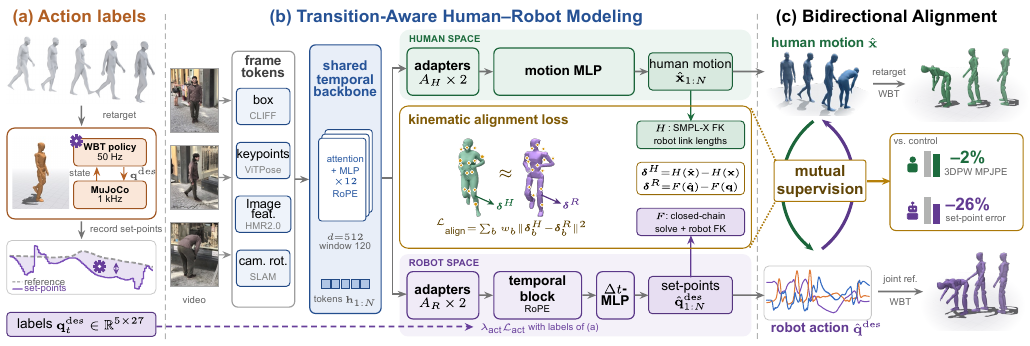}
\vspace{-1.6em}
\caption{
\textbf{Overview of \method.}
(a)~Each training motion is executed once by the (WBT) controller in closed-loop simulation, yielding set-point executed humanoid action labels.
(b)~Transition-aware human--robot modeling: a shared temporal backbone feeds human and action branches with separate adapters.
(c)~Bidirectional alignment: alternating teacher roles enable mutual supervision, improving both branches over a control without the new modules (\cref{sec:exp_ablation}).
At inference, either output drives the same controller.
}
\label{fig:overview}
\vspace{-0.8em}
\end{figure*}

\paragraph{Human motion recovery.}
Regressing SMPL~\cite{loper2015smpl,pavlakos2019smplx} bodies from images, by optimization~\cite{bogo2016smplify} or regression~\cite{kanazawa2018hmr,kolotouros2019spin,kocabas2021pare,zhang2021pymaf,lin2021metro,li2022cliff,li2021hybrik,goel2023hmr2,wang2023refit,dwivedi2024tokenhmr,baradel2024multihmr,patel2025camerahmr,sarandi2024nlf,wang2025prompthmr}, and from videos~\cite{kanazawa2019hmmr,kocabas2020vibe,choi2021tcmr,zhu2023motionbert} has developed into recovering motion in a gravity-aware world frame from moving cameras, by combining SLAM~\cite{teed2021droid,teed2023dpvo} with learned motion priors~\cite{rempe2021humor,ye2023slahmr,kocabas2024pace} or by regressing the root trajectory directly~\cite{yuan2022glamr,li2022dnd,sun2023trace,shin2024wham,wang2024tram,yin2024whac}.
GVHMR~\cite{shen2024gvhmr} predicts orientations in a per-frame gravity-view frame, GENMO~\cite{li2025genmo} unifies estimation and generation in one diffusion Transformer, DuoMo~\cite{wang2026duomo} couples camera- and world-space motion diffusion, and diffusion priors add robustness~\cite{li2024coin,zhang2024rohm}.
While these methods focus on visual reconstruction, \method jointly learns human motion recovery and robot action prediction in a single training stage, incorporating dynamics transition to improve both reconstruction fidelity and robot executability.

\paragraph{Physically plausible motion recovery and generation.}
Prior work improves motion plausibility through trajectory optimization under dynamics and contact constraints~\cite{shimada2020physcap,shimada2021neuralphyscap,xie2021physics,gartner2022trajopt,gartner2022diffphy}, explicit balance and ground-contact objectives~\cite{tripathi2023ipman}, physics-based character control~\cite{yuan2020rfc,yuan2021simpoe,luo2021kinpoly,luo2022embodied}, and learned motion refinement~\cite{zhang2024physpt}.
More recently, FlowHMR and ViT-H~\cite{wang2026flowhmr,he2026vith} improve the trackability of video-conditioned motion through reward- or preference-based fine-tuning using simulated execution feedback.
For text-conditioned generation, PhysDiff~\cite{yuan2023physdiff} incorporates physics-based motion projection into diffusion sampling, while RLPF and GenTrack~\cite{yue2025rlpf,ling2026gentrack} fine-tune motion generators using tracking feedback, with GenTrack additionally updating the tracking controller.
\method learns human motion recovery and robot action prediction within a single model and training stage, using executed actions as direct supervision for transition dynamics and bidirectional kinematic alignment to connect reconstruction fidelity with robot executability.

\paragraph{Humanoid motion imitation from human data.}
Humanoid imitation pipelines typically retarget human motion into reference trajectories and execute them with learned whole-body tracking~\cite{araujo2025gmr,he2025hover,he2024omnih2o,ze2025twist,fu2024humanplus,ji2024exbody2,he2025asap,liao2025beyondmimic,mao2024uh1,allshire2025videomimic}.
Successful imitation depends on both the physical feasibility of these references and the controller's ability to track them.
To improve execution, H2O~\cite{he2024h2o} filters out motions that a privileged policy cannot track, while ExBody~\cite{cheng2024exbody} combines upper-body imitation with velocity-conditioned locomotion.
\method addresses executability via motion recovery itself by jointly learning human motion recovery and robot action prediction in single stage, using executed-action supervision to capture transition dynamics.

\section{\method}\label{sec:method}

Given a monocular video, \method recovers world-grounded human motion and corresponding actions for humanoid execution (\cref{fig:overview}).
The overall framework is trained using executed-action labels collected offline through controller simulation (\cref{sec:labels}).
On a shared temporal backbone, we introduce three components: task-specific branches decoding human motion and robot actions from shared tokens, temporal action modeling that predicts controller set-points from surrounding frames (\cref{sec:model}), and bidirectional kinematic alignment enabling mutual supervision during training (\cref{sec:align}).
We also detail the training objective and two output formats (\cref{sec:impl}).

\subsection{Problem Formulation and Framework}
\label{sec:labels}

\paragraph{Problem formulation.} \method aims to recover, for every frame $t$, both the motion of the person and the action of the humanoid robot. 
Following GVHMR and GENMO~\cite{shen2024gvhmr,li2025genmo}, the human motion is $\mathbf{x}_t = (\Gamma^{gv}_t, \mathbf{v}^{root}_t, \bm\theta_t, \bm\beta_t, \mathbf{t}^{c}_t, \mathbf{c}_t)$: the body orientation in the gravity-view frame, the root velocity in the body frame, the SMPL-X~\cite{pavlakos2019smplx} pose and shape, the camera-frame root translation and the contact probabilities of hands and feet; integrating the root velocities in the world frame yields the trajectory. 
The action $\qdes_t\in\R^{K\times 27}$ consists of the joint set-points that the controller of a humanoid with 27 actuated joints (LimX Luna~\cite{limx_luna}) applies between frames $t$ and $t{+}1$, at $K{=}5$ sub-intervals. 
Whereas $\mathbf{x}_t$ describes a pose, $\qdes_t$ describes how the robot gets from one pose to the next, and thus makes the inter-frame dynamics an explicit regression target.

\paragraph{Framework.} To support human motion recovery and humanoid imitation within one model, \method encodes the video into shared tokens with a temporal Transformer and decodes $\hat{\mathbf{x}}_{1:N}$ and $\hq_{1:N}$ through task-specific branches with separate residual adapters (\cref{sec:model}).
During training, each branch receives its own label supervision, while bidirectional kinematic alignment compares their label-centred residuals in common body frames, allowing the two tasks to guide each other (\cref{sec:align}).
For training data, human motion labels are provided in original datasets, whereas executed-action labels are generated offline through closed-loop simulation of deployed trackers.
At inference, the two predictions support human motion recovery and humanoid imitation action signal in a single forward pass (\cref{sec:impl}).

\paragraph{Executed humanoid action.}
We generate dense action labels offline by retargeting each training motion with GMR~\cite{araujo2025gmr} and executing the resulting reference in MuJoCo~\cite{todorov2012mujoco} under a fixed whole-body tracking (WBT) policy~\cite{fu2026cosa05}.
At each control update, the policy takes proprioceptive observations and future reference configurations as input and outputs joint set-points for a low-level proportional-derivative (PD) controller.
To align these commands with the human motion sequence, we divide the interval between consecutive timestamps $\tau_t$ and $\tau_{t+1}$ into $K$ sub-intervals of duration $\Delta t_t=(\tau_{t+1}-\tau_t)/K$ and assign each sub-interval the set-point tracked by the PD controller during it.
Policy updates follow the controller's native timing and are aligned to the nearest sub-interval boundary, with set-points held constant between updates.
Execution proceeds in closed loop, carrying both the simulated robot state and controller state across successive frames, so the recorded commands reflect feedback adjustments for motion tracking, balance, and contact.
The resulting labels describe the commands used to execute each transition, distinct from both the retargeted reference configurations and the joint trajectories actually reached by the robot.
These labels provide temporally aligned supervision for learning the controller's action requirements for tracking, balance, and contact.

\subsection{Transition-Aware Human--Robot Modeling}
\label{sec:model}

\method connects human motion recovery and robot action prediction through complementary representations of the same motion transitions.
To learn this connection, a shared temporal backbone encodes motion context across frames, and task-specific adapters refine its features for human and action decoding.
The action branch further models temporal dependencies to predict a sequence of controller commands within each transition.
Supervision from both branches shapes the shared representation, allowing reconstruction cues and transition-dependent execution requirements to jointly guide learning.

\paragraph{Shared temporal backbone.}
To provide both tasks with common motion context, we form $\mathbf{f}_t\in\R^d$ by summing embeddings of CLIFF bounding-box parameters~\cite{li2022cliff}, ViTPose keypoints~\cite{xu2022vitpose}, frozen image features~\cite{goel2023hmr2}, and relative camera rotation.
Missing conditions are flagged, while motion-capture sequences without video use synthetic image-plane inputs rendered with static and moving cameras~\cite{shin2024wham,shen2024gvhmr}.
A temporal Transformer~\cite{vaswani2017attention} with rotary position embeddings~\cite{su2024roformer} maps $\mathbf{f}_{1:N}$ to shared tokens $\mathbf{h}_{1:N}$ using bounded temporal windows during training and inference.
These tokens aggregate observations for human reconstruction and transition-aware action prediction.

\paragraph{Task-specific adapters.}
Human reconstruction and controller action prediction require distinct features, so we equip each branch with residual bottleneck adapters~\cite{rebuffi2017adapters,houlsby2019adapters}.
Each adapter takes the form:
\begin{equation}
  A(\mathbf{h}) = \mathbf{h} + W_{\uparrow}\,\mathrm{SiLU}\big(W_{\downarrow}\,\mathrm{LN}(\mathbf{h})\big),
\end{equation}
where $W_{\downarrow}\in\R^{r\times d}$ and $W_{\uparrow}\in\R^{d\times r}$ project through a bottleneck of width $r<d$.
The zero-initialized output projection initially preserves shared features, allowing each branch to learn task-specific residual transformations.
The human branch applies GENMO's regression MLP~\cite{li2025genmo} to $A_H(\mathbf{h}_t)$ to decode $\hat{\mathbf{x}}_t$, comprising body pose, global orientations, body shape, and root velocity.
Camera parameters and contacts are decoded directly from $\mathbf{h}_t$, with contacts correcting the root trajectory at inference as in GVHMR~\cite{shen2024gvhmr}.
The action branch passes $A_R(\mathbf{h}_{1:N})$ to a temporal module that predicts commands throughout the corresponding transitions.

\paragraph{Temporal action branch.}
Humanoid actions depend on upcoming reference configurations and the state accumulated during preceding execution, requiring temporal context beyond the current pose.
To capture these dependencies, we process projected action tokens with bidirectional Transformer blocks using rotary position embeddings and final layer normalization, collectively denoted by $\mathcal{T}_R$.
A per-frame MLP combines these features with a duration embedding to predict $K$ joint set-points:
\begin{equation}
  \hq_t = \mu + \sigma\odot \mathrm{MLP}\big(\mathcal{T}_R(A_R(\mathbf{h}_{1:N}))_t \oplus \phi(\Delta t_t)\big),
\end{equation}
where $\oplus$ denotes concatenation, $\phi$ embeds $\Delta t_t$, and $\mu,\sigma$ are per-joint label statistics.
These statistics normalize action targets to reduce scale differences between joints and restore predictions to their original units.
Duration conditioning makes transition timing explicit, allowing the predictor to account for different video frame rates.
Executed-action supervision propagates through the temporal module and task-specific adapters into the shared backbone.
Human motion recovery thus learns from reconstruction targets and transition-dependent control requirements, coupling recovery fidelity with robot executability.

\begin{figure}[t]
\centering
\input{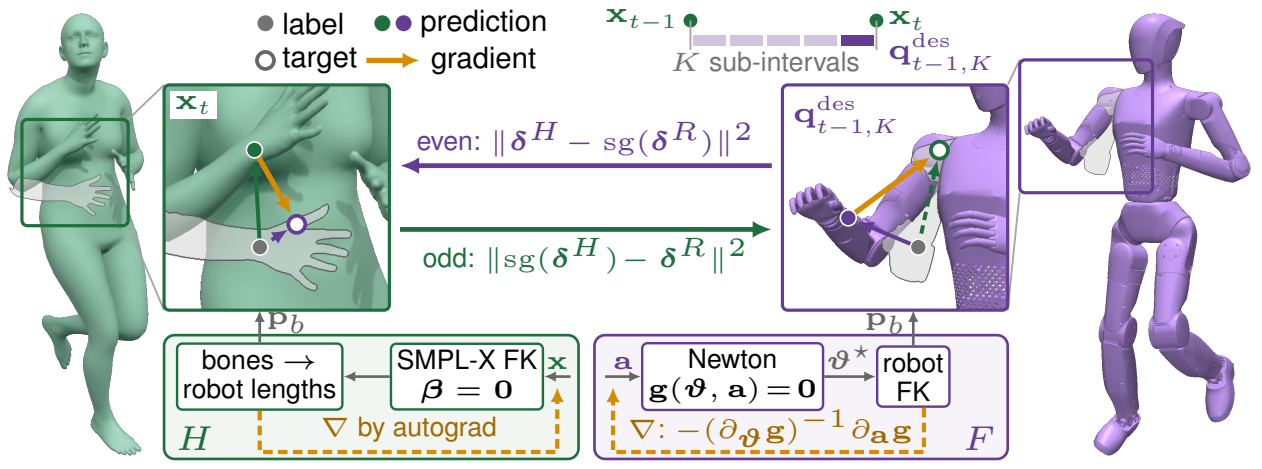}
\vspace{-1.8em}
\caption{\textbf{Bidirectional alignment in detail} (right wrist, illustrative predictions). $H$, $F$ map labels and predictions alike; each step pulls one branch toward its own label plus the other's frozen residual, back-propagating through $H$ or implicitly through $F$.}
\label{fig:align}
\vspace{-1em}
\end{figure}

\subsection{Bidirectional Kinematic Alignment}
\label{sec:align}

To unify both branches around faithful reconstruction and reliable execution of the same underlying motion trajectory, we introduce bidirectional kinematic alignment (Fig.~\ref{fig:align}).
We map predictions and labels to common body positions and align label-relative residuals, alternating teacher roles to enable mutual supervision beyond shared features.

\paragraph{Mapping outputs to common body positions.}
To compare the two representations of the same motion, we map predictions and labels into a common geometric space using fixed, differentiable maps $H$ and $F$ (Fig.~\ref{fig:align}).
These maps yield root-local positions $\mathbf{p}_b$ for 14 corresponding bodies: the hips, knees, ankles, waist, neck, shoulders, elbows, and wrists.
To account for different body proportions, $H$ applies SMPL-X forward kinematics with neutral shape and rescales each bone to the corresponding robot-link length while preserving its direction.
For robot actions, $F$ solves the linkage closure constraints $\mathbf{g}(\bm\vartheta,\mathbf{a})=\mathbf{0}$ to convert actuator set-points $\mathbf{a}$ into serial-chain joint angles $\bm\vartheta$, then applies robot forward kinematics.
To propagate alignment gradients, we differentiate $H$ automatically and implicitly differentiate the closure solution $\bm\vartheta^\star$ obtained by Newton iterations:
\begin{equation}
  \frac{\partial\bm\vartheta^\star}{\partial\mathbf{a}}
  =
  -\Big(\frac{\partial\mathbf{g}}{\partial\bm\vartheta}\Big)^{-1}
  \frac{\partial\mathbf{g}}{\partial\mathbf{a}}.
\end{equation}
Frames with invalid predicted or target linkage closure retain action supervision but are excluded from alignment.

\paragraph{Aligning prediction errors.}
Retargeting and controller corrections give the two branches different reference configurations even for the same underlying motion.
To preserve their respective supervision, we therefore align label-relative residuals rather than absolute positions.
To compare temporally corresponding quantities, we pair human frame $t$ with the final set-point of the preceding transition interval:
\begin{align}
  \bm\delta^{H}_{t,b}
  &=
  \mathbf{p}_b\big(H(\hat{\mathbf{x}}_t)\big)
  -
  \mathbf{p}_b\big(H(\mathbf{x}_t)\big), \\
  \bm\delta^{R}_{t,b}
  &=
  \mathbf{p}_b\big(F(\hq_{t-1,K})\big)
  -
  \mathbf{p}_b\big(F(\qdes_{t-1,K})\big), \\
  \cL_{\text{align}}
  &=
  \frac{1}{|\mathcal{S}|}
  \sum_{t\in\mathcal{S}}\sum_b
  w_b\big\|\bm\delta^{H}_{t,b}-\bm\delta^{R}_{t,b}\big\|_2^2,
\end{align}
where $w_b$ are normalized retargeting position weights and $\mathcal{S}$ contains sampled frames with valid human and action labels and feasible linkage closure.
Exact predictions yield zero alignment loss despite reference differences, while matching nonzero residuals remain penalized by the respective supervised losses.
To enable mutual guidance, we alternate teacher roles (Fig.~\ref{fig:align}): the robot branch supplies $\mathrm{sg}(\bm\delta^R)$ on even steps, and the human branch supplies $\mathrm{sg}(\bm\delta^H)$ on odd steps, where $\mathrm{sg}$ denotes stop-gradient.
Alignment gradients flow only through the student residual, guiding its mapped positions toward its own label positions plus the teacher's frozen residual; both label losses remain active.
To introduce this guidance gradually, we increase the alignment weight linearly to $\lambda_{\text{align}}$ after warm-up.
This bidirectional supervision unifies recovery fidelity and robot executability around the same underlying motion within a single training stage.
Alignment requires no simulation or additional trainable parameters and is omitted at inference.
\subsection{Training and Inference}
\label{sec:impl}

\paragraph{Training objective.} To train both branches jointly, the human branch is supervised with the estimation losses $\cL_{\text{human}}$ of GVHMR and GENMO (motion vector, camera- and world-frame joints, vertices and translations, 2D reprojection and contacts, here with an additional joint-velocity and smoothness term) and with $\cL_{\text{foot}}$, the squared error of the world velocity of the ankle and foot joints on frames where the ground-truth foot is in contact (speed below 0.15\,m/s), which also corrects the root motion that causes skating. The action branch is trained with the masked MSE $\cL_{\text{act}}\propto\sum_{t} m_{t} \| (\hq_{t}-\qdes_{t})/\sigma \|_2^2$. 
The total objective is:
\begin{equation}
  \cL=\cL_{\text{human}}+\lambda_{\text{foot}}\cL_{\text{foot}}+\lambda_{\text{act}}\cL_{\text{act}}+\lambda_{\text{align}}\cL_{\text{align}},
\end{equation}
with $\lambda_{\text{foot}}{=}5$, $\lambda_{\text{act}}{=}1$ and $\lambda_{\text{align}}{=}0.5$. \method is trained with AdamW~\cite{loshchilov2019adamw}, a learning rate of $10^{-4}$ and a global batch of 2048 windows on 16 accelerators; we report the checkpoint after \numActStep\ steps.

\paragraph{Human motion recovery and humanoid imitation.}
At inference, \method recovers human motion and predicts robot actions in a single forward pass, without test-time simulation or motion optimization.
The human output provides SMPL-X body motion and a world-grounded trajectory, while the action output supports humanoid imitation through a fixed WBT controller.
To construct the controller references, we convert predicted actuator set-points into joint configurations by solving the linkage constraints and combine them with the root trajectory retargeted from the human output.
These joint references retain the sub-interval timing of the action predictions, describing the progression of each motion transition.
The WBT controller tracks these references in closed loop and generates its own low-level PD targets, allowing the predicted actions to guide execution while preserving feedback control.

\section{Experiments}
\label{sec:experiments}
\input{tables/numbers_ablation_modules}

\begin{table*}[t]
\centering
\setlength{\tabcolsep}{3.4pt}
\footnotesize
\renewcommand{\arraystretch}{0.9}
\caption{\textbf{Camera-space human motion recovery.}
Errors are in mm and Accel in m/s$^2$; joint counts are in parentheses.
$^*$Trained with 3DPW, including \method.
We evaluate \method\ with the official GENMO/GVHMR code and decode camera-frame translation using dataset intrinsics.
$^\ddagger$Our evaluation without flip augmentation.}
\label{tab:cam}
\resizebox{\textwidth}{!}{%
\begin{tabular}{ll cccc cccc cccc}
\toprule
 & & \multicolumn{4}{c}{3DPW (14)} & \multicolumn{4}{c}{RICH (24)} & \multicolumn{4}{c}{EMDB-1 (24)} \\
\cmidrule(lr){3-6}\cmidrule(lr){7-10}\cmidrule(lr){11-14}
 & Method & PA-MPJPE & MPJPE & PVE & Accel & PA-MPJPE & MPJPE & PVE & Accel & PA-MPJPE & MPJPE & PVE & Accel \\
\midrule
\multirow{7}{*}{\rotatebox{90}{\scriptsize per-frame}} & CLIFF$^*$~\cite{li2022cliff} & 43.0 & 69.0 & 81.2 & 22.5 & 56.6 & 102.6 & 115.0 & 22.4 & 68.1 & 103.3 & 128.0 & 24.5 \\
 & HybrIK$^*$~\cite{li2021hybrik} & 41.8 & 71.6 & 82.3 & -- & 56.4 & 96.8 & 110.4 & -- & 65.6 & 103.0 & 122.2 & -- \\
 & HMR2.0~\cite{goel2023hmr2} & 44.4 & 69.8 & 82.2 & 18.1 & 48.1 & 96.0 & 110.9 & 18.8 & 60.6 & 98.0 & 120.3 & 19.8 \\
 & ReFit$^*$~\cite{wang2023refit} & 40.5 & 65.3 & 75.1 & 18.5 & 47.9 & 80.7 & 92.9 & 17.1 & 58.6 & 88.0 & 104.5 & 20.7 \\
 & CameraHMR$^*$~\cite{patel2025camerahmr} & 35.1 & 56.0 & 65.9 & -- & \best{34.0} & \second{55.7} & \second{64.4} & -- & 43.3 & 70.2 & 81.7 & -- \\
 & NLF~\cite{sarandi2024nlf} & 37.3 & 60.3 & 71.4 & -- & -- & -- & -- & -- & 41.2 & 69.6 & 82.4 & -- \\
 & PromptHMR$^*$~\cite{wang2025prompthmr} & 36.6 & 58.7 & 69.4 & -- & 37.3 & 56.6 & 65.5 & -- & \second{41.0} & 71.7 & 84.5 & -- \\
\midrule
\multirow{10}{*}{\rotatebox{90}{\scriptsize temporal}} & VIBE$^*$~\cite{kocabas2020vibe} & 51.9 & 82.9 & 98.4 & 18.5 & 68.4 & 120.5 & 140.2 & 21.8 & 81.4 & 125.9 & 146.8 & 26.6 \\
 & TRACE$^*$~\cite{sun2023trace} & 50.9 & 79.1 & 95.4 & 28.6 & -- & -- & -- & -- & 70.9 & 109.9 & 127.4 & 25.5 \\
 & SLAHMR~\cite{ye2023slahmr} & 55.9 & -- & -- & -- & 52.5 & -- & -- & 9.4 & 69.5 & 93.5 & 110.7 & 7.1 \\
 & PACE~\cite{kocabas2024pace} & -- & -- & -- & -- & 49.3 & -- & -- & 8.8 & -- & -- & -- & -- \\
 & WHAM$^*$~\cite{shin2024wham} & 35.9 & 57.8 & 68.7 & 6.6 & 44.3 & 80.0 & 91.2 & 5.3 & 50.4 & 79.7 & 94.4 & 5.3 \\
 & GVHMR$^*$~\cite{shen2024gvhmr} & 36.2 & 55.6 & 67.2 & 5.0 & 39.5 & 66.0 & 74.4 & \second{4.1} & 42.7 & 72.6 & 84.2 & 3.6 \\
 & TRAM$^*$~\cite{wang2024tram} & 35.6 & 59.3 & 69.6 & 4.9 & -- & -- & -- & -- & 45.7 & 74.4 & 86.6 & 4.9 \\
 & GENMO$^*$~\cite{li2025genmo} & \second{34.6} & \second{53.9} & \second{65.8} & 5.2 & 39.1 & 66.8 & 75.4 & \second{4.1} & 42.5 & 73.0 & 84.8 & 3.8 \\
 & DuoMo$^*$~\cite{wang2026duomo} & 36.3$^\ddagger$ & 57.7$^\ddagger$ & 69.5$^\ddagger$ & 7.1$^\ddagger$ & \second{34.8} & \best{51.4} & \best{58.9} & -- & 41.8 & \second{67.1} & \second{78.2} & -- \\
 & FlowHMR~\cite{wang2026flowhmr}$^\ddagger$ & 42.8 & 72.9 & 85.7 & \second{4.8} & -- & -- & -- & -- & 49.4 & 92.5 & 105.2 & \second{3.1} \\
\midrule
\rowcolor{red!6} & \method & \best{34.3} & \best{53.7} & \best{64.8} & \best{4.7} & 38.7 & 65.9 & 73.8 & \best{4.0} & \best{40.8} & \best{64.7} & \best{76.3} & \best{2.9} \\
\bottomrule
\end{tabular}}
\vspace{-0.8em}
\end{table*}

We evaluate the human motion recovered by \method on standard benchmarks (\cref{sec:exp_human}), its execution by a humanoid (\cref{sec:exp_robot}), the contribution of its components (\cref{sec:exp_ablation}) and its deployment on the real robot (\cref{sec:exp_real}).

\subsection{Setup}
\label{sec:exp_setup}

\paragraph{Benchmarks and metrics.} 
We evaluate on 3DPW, EMDB and RICH~\cite{vonmarcard2018_3dpw,kaufmann2023emdb,huang2022rich} with the official GENMO/GVHMR test code and report PA-MPJPE, MPJPE, PVE (mm) and Accel (m/s$^2$) in camera coordinates and WA-/W-MPJPE$_{100}$ (mm), RTE (\%), jitter (m/s$^3$) and foot sliding (FS, mm) in world coordinates, with FlipEval on 3DPW and EMDB as in GENMO~\cite{li2025genmo}. 
\method receives the camera rotation of the DROID-SLAM trajectories released with the GVHMR test sets; in \cref{tab:world}, its camera-space prediction is placed in the world with the metric trajectory, as TRAM does. 
On EMDB, we keep the network input $f{=}\max(w,h)$ of GENMO's protocol but decode the camera-frame translation with the dataset intrinsics.

\paragraph{Compared methods and robot benchmark.} 
Tabs.~\ref{tab:cam} and~\ref{tab:world} report published results for the methods compared in GENMO~\cite{li2025genmo} and DuoMo~\cite{wang2026duomo}.
For the robot benchmark, we run the released models of WHAM~\cite{shin2024wham}, GVHMR~\cite{shen2024gvhmr}, TRAM~\cite{wang2024tram}, GENMO, DuoMo, and FlowHMR~\cite{wang2026flowhmr} on a common test set comprising all 3DPW and EMDB test videos and RICH videos with available raw frames.
We split the recovered world-frame motions into a total of \nSegAll\ segments of up to 10\,s and apply the same root-trajectory correction across methods to remove vertical drift and foot sliding during contact.
Each segment is retargeted with GMR and executed at 50\,Hz using the same fixed WBT controller and simulator used for label generation.
Ground-truth execution identifies the \nSegFeas\ GT-feasible segments; the primary metric is success rate, defined as the fraction of segments executed without a fall.

\begin{table}[t]
\centering
\setlength{\tabcolsep}{3.0pt}
\footnotesize
\renewcommand{\arraystretch}{0.9}
\caption{\textbf{World-grounded human motion} on EMDB-2 (24 joints; WA/W-MPJPE$_{100}$ and FS in mm, RTE in \%, Jitter in m/s$^3$). Top: published; $^\ddagger$our run. \method's camera-space prediction is placed in the world with the DROID-SLAM trajectory and TRAM's ($^\star$) or its own ($^\diamond$) translation.}
\label{tab:world}
\resizebox{\columnwidth}{!}{%
\begin{tabular}{l ccccc}
\toprule
Method & WA-MPJPE$_{100}$ & W-MPJPE$_{100}$ & RTE & Jitter & FS \\
\midrule
GLAMR~\cite{yuan2022glamr} & 280.8 & 726.6 & 11.4 & 46.3 & 20.7 \\
TRACE~\cite{sun2023trace} & 529.0 & 1702.3 & 17.7 & 2987.6 & 370.7 \\
SLAHMR~\cite{ye2023slahmr} & 326.9 & 776.1 & 10.2 & 31.3 & 14.5 \\
COIN~\cite{li2024coin} & 152.8 & 407.3 & 3.5 & -- & -- \\
WHAM~\cite{shin2024wham} (DPVO) & 135.6 & 354.8 & 6.0 & 22.5 & 4.4 \\
GVHMR~\cite{shen2024gvhmr} (DPVO) & 111.0 & 276.5 & 2.0 & 16.7 & \second{3.5} \\
TRAM~\cite{wang2024tram} (DROID-SLAM) & 76.4 & 222.4 & \second{1.4} & 18.1 & 11.0 \\
GENMO~\cite{li2025genmo} (DROID-SLAM) & 74.3 & 202.1 & \best{1.2} & 17.8 & 8.8 \\
DuoMo~\cite{wang2026duomo} & \second{67.6} & \best{171.6} & \best{1.2} & 8.7 & 3.7 \\
FlowHMR~\cite{wang2026flowhmr}$^\ddagger$ & 107.3 & 273.6 & 2.4 & \best{5.4} & \best{2.9} \\
\midrule
\rowcolor{red!6}\method (DROID-SLAM)$^\diamond$ & \best{63.4} & \second{176.7} & \second{1.4} & \second{7.2} & 6.6 \\
\rowcolor{red!6}\method (DROID-SLAM)$^\star$ & 68.2 & 185.7 & \second{1.4} & 7.3 & 7.6 \\
\bottomrule
\end{tabular}}
\vspace{-1.2em}
\end{table}

\begin{figure*}[t]
\centering
\includegraphics[width=\textwidth]{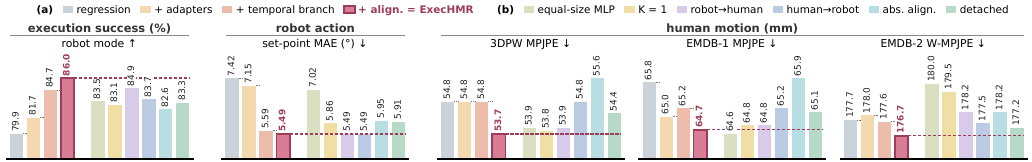}
\vspace{-1.6em}
\caption{\textbf{Module ablation under executed-action supervision.} All variants share the data, losses, optimizer, budget and seed of \method and change only the listed component. (a)~Components added one at a time to simple action regression, a per-frame action MLP on the shared tokens; (b)~single changes to \method (dashed): temporal blocks replaced by an equal-size per-frame MLP, one set-point per transition ($K{=}1$), alignment in one direction only (robot$\to$human, human$\to$robot) or of absolute positions, and detached actions without alignment. Execution success is evaluated in robot mode; set-point MAE on the training stream (45k--50k steps); EMDB-2: W-MPJPE$_{100}$ of the predicted world trajectory.}
\label{fig:ablation_main}
\vspace{-0.9em}
\end{figure*}

\subsection{Human Motion Recovery}
\label{sec:exp_human}

\paragraph{Camera space (\cref{tab:cam}).}
On 3DPW, \method achieves the lowest PA-MPJPE (34.3\,mm), MPJPE (53.7\,mm), and PVE (64.8\,mm) among the compared methods.
It also ranks first on all three metrics on EMDB-1, reducing MPJPE by 2.4\,mm and PVE by 1.9\,mm relative to DuoMo.
On RICH, \method improves on GVHMR and GENMO across these three metrics, although CameraHMR, PromptHMR, and DuoMo achieve lower MPJPE.
\method also achieves the lowest reported Accel on each dataset (4.7, 4.0, and 2.9\,m/s$^2$ on 3DPW, RICH, and EMDB-1, respectively).
With a shared temporal backbone for human motion recovery and robot action prediction, \method outperforms GVHMR and GENMO on all four metrics averaged across datasets.
Its mean MPJPE is 61.4\,mm, compared with 64.7\,mm for GVHMR and 64.6\,mm for GENMO.

\paragraph{World space (\cref{tab:world}).}
Using the metric DROID-SLAM trajectory and its own camera-frame translation, \method achieves the lowest WA-MPJPE$_{100}$ on EMDB-2 (63.4\,mm versus DuoMo's 67.6\,mm and GENMO's 74.3\,mm).
It ranks second in W-MPJPE$_{100}$ (176.7\,mm), behind DuoMo (171.6\,mm) and ahead of GENMO (202.1\,mm).
With the trajectory fixed, substituting TRAM's translation raises these errors to 68.2 and 185.7\,mm, respectively.
During training, the human motion and robot action branches alternate teacher roles to match their label-relative position residuals.
This bidirectional kinematic alignment modestly reduces world-space joint error (\cref{sec:exp_ablation}).

\subsection{Robot Executability}
\label{sec:exp_robot}

\begin{table}[t]
\centering
\setlength{\tabcolsep}{2.6pt}
\footnotesize
\renewcommand{\arraystretch}{0.9}
\caption{\textbf{Robot benchmark} (\nSegAll\ segments of the test videos on which every method can be run; ground-fixed, retargeted with GMR). \method executes its predicted set-points. Succ.: fall-free segments, all and GT-feasible. Track/Fid.: joint error (rad) to own reference/ground truth; Slip: foot slip (cm/s). $^\dagger$GT gravity.}
\label{tab:robot}
\resizebox{\columnwidth}{!}{%
\begin{tabular}{l cc cccc ccc}
\toprule
 & \multicolumn{2}{c}{Succ.\,(\%)\,$\uparrow$}
 & \multicolumn{4}{c}{Succ. per dataset (\%)\,$\uparrow$}
 & Track & FID & Slip \\
\cmidrule(lr){2-3}\cmidrule(lr){4-7}
Method & all & feas. & 3DPW & EMDB-1 & EMDB-2 & RICH
 & $\downarrow$ & $\downarrow$ & $\downarrow$ \\
\midrule
\textcolor{gray}{GT motion (oracle)}
 & \textcolor{gray}{77.7}
 & \textcolor{gray}{100.0}
 & \textcolor{gray}{79.9}
 & \textcolor{gray}{71.1}
 & \textcolor{gray}{80.7}
 & \textcolor{gray}{74.6}
 & \textcolor{gray}{0.089}
 & \textcolor{gray}{0.089}
 & \textcolor{gray}{19.3} \\
\midrule
WHAM~\cite{shin2024wham}$^\dagger$
 & 73.8 & 82.6 & 82.7 & 69.9 & 72.0 & 64.8
 & 0.079 & 0.172 & 17.0 \\
GVHMR~\cite{shen2024gvhmr}
 & 79.0 & \second{89.2} & 87.1 & \best{77.1} & \second{76.7} & 70.4
 & \second{0.077} & \second{0.165} & 14.4 \\
TRAM~\cite{wang2024tram}$^\dagger$
 & 75.6 & 87.8 & 82.0 & 73.5 & 73.3 & 70.4
 & 0.084 & 0.181 & 17.3 \\
DuoMo~\cite{wang2026duomo}$^\dagger$
 & 76.1 & 86.6 & 86.3 & 72.3 & 71.3 & 70.4
 & 0.087 & 0.168 & 16.8 \\
GENMO~\cite{li2025genmo}
 & \second{79.7} & \second{89.2} & \second{87.8} & \best{77.1} & 75.3 & \best{76.1}
 & \best{0.076} & 0.168 & \second{14.3} \\
FlowHMR~\cite{wang2026flowhmr}
 & 76.3 & 86.0 & 83.5 & \best{77.1} & 71.3 & \second{71.8}
 & 0.081 & 0.169 & 16.3 \\
\midrule
\rowcolor{red!6}\method
 & \best{86.0} & \best{95.6} & \best{96.4} & \second{75.9} & \best{90.0} & 69.0
 & 0.078 & \best{0.162} & \best{10.8} \\
\bottomrule
\end{tabular}}
\vspace{-1.2em}
\end{table}

\begin{figure}[t]
\centering
\includegraphics[width=\linewidth]{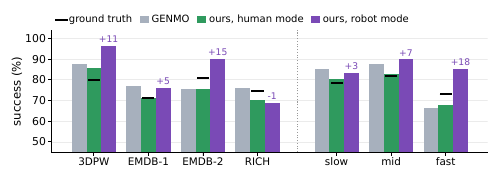}
\vspace{-1.8em}
\caption{\textbf{Execution success across motion speeds.} Success rates of GENMO and both modes of \method, grouped by dataset and joint-speed tercile. Ticks indicate ground-truth success; numbers indicate robot-minus-human differences in percentage points.}
\label{fig:robotmode}
\vspace{-1em}
\end{figure}
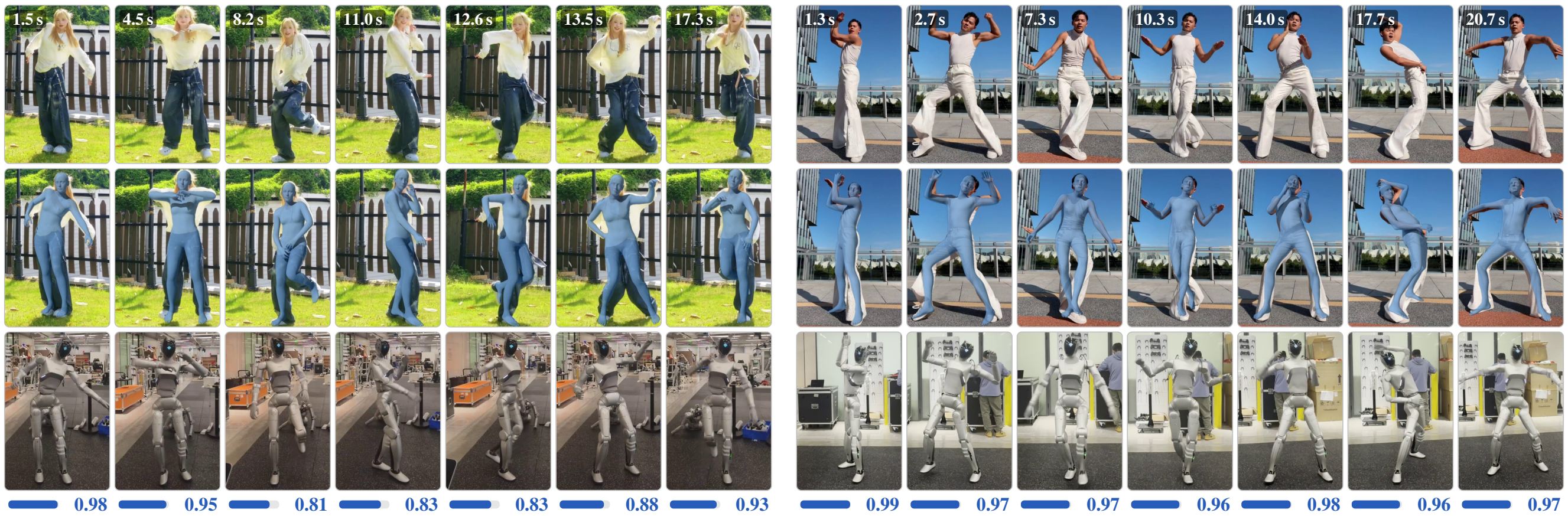
\begin{figure*}[t]
\centering
\definecolor{rrOurs}{RGB}{40,92,186}
\resizebox{\textwidth}{!}{%
\begin{tikzpicture}[x=1pt,y=1pt,font=\scriptsize]
\def\tw{33.5}\def\th{50.45}\def\tg{1.6}\def\rg{1.6}\def\gg{8}\def\bw{16}
\def\tile#1#2#3{
  \begin{scope}
    \clip[rounded corners=1.5pt] (#1,#2) rectangle ++(\tw,\th);
    \node[anchor=south west,inner sep=0pt] at (#1,#2) {\includegraphics[width=\tw pt,height=\th pt]{#3}};
  \end{scope}
  \draw[rounded corners=1.5pt,draw=black!25,line width=0.3pt] (#1,#2) rectangle ++(\tw,\th);}%
\pgfmathsetmacro\ym{\th+\rg}\pgfmathsetmacro\yv{2*(\th+\rg)}%
\foreach \c/\vid/\idx/\t/\s in {1/a/1/1.5/0.98,2/a/2/4.5/0.95,3/a/3/8.2/0.81,4/a/4/11.0/0.83,5/a/5/12.6/0.83,6/a/6/13.5/0.88,
    7/a/7/17.3/0.93,8/b/1/1.3/0.99,9/b/2/2.7/0.97,10/b/3/7.3/0.97,11/b/4/10.3/0.96,12/b/5/14.0/0.98,13/b/6/17.7/0.96,
    14/b/7/20.7/0.97}{
  \pgfmathsetmacro\tx{(\c-1)*(\tw+\tg)+(\c>7)*(\gg-\tg)}
  \tile{\tx}{\yv}{figures/real/\vid_in\idx.jpg}
  \tile{\tx}{\ym}{figures/real/\vid_mesh\idx.jpg}
  \tile{\tx}{0}{figures/real/\vid_robot\idx.jpg}
  \node[anchor=north west,inner sep=1pt,fill=black,fill opacity=0.55,text opacity=1,text=white,rounded corners=1pt,
        font=\fontsize{5.4}{6}\selectfont\bfseries] at (\tx+1.3,\yv+\th-1.3) {\t\,s};
  \fill[black!10,rounded corners=1.1pt] (\tx+1.2,-5.8) rectangle ++(\bw,2.6);
  \fill[rrOurs,rounded corners=1.1pt] (\tx+1.2,-5.8) rectangle ++(\bw*\s,2.6);
  \node[anchor=east,inner sep=0pt,text=rrOurs,font=\fontsize{6}{6.6}\selectfont\bfseries] at (\tx+\tw-0.6,-4.5) {\s};}
\end{tikzpicture}}
\vspace{-1.6em}
\caption{\textbf{Real-robot deployment} on two in-the-wild dance videos (left, right). Rows: input video; human mesh recovered by \method; real LimX Luna humanoid executing the recovered motion, at the same instants. Bars: pose similarity of dancer and robot, the mean cosine of the differences between their limb angles relative to the torso in ViTPose keypoints.}
\label{fig:real}
\vspace{-0.9em}
\end{figure*}

\paragraph{Execution success.}
Without test-time joint smoothing, \method in robot mode achieves the highest execution success among the compared methods, reaching \sAllOursAct\% overall and \sFeasOursAct\% on GT-feasible segments (\cref{tab:robot}).
These results surpass the strongest baseline by 6.3 and 6.4 percentage points, respectively.
Paired sign tests further confirm significant execution gains over every compared method.
On 3DPW and EMDB-2, success rates reach 96.4\% and 90.0\%, exceeding ground-truth execution by 16.5 and 9.3 percentage points.
Beyond fall-free execution, \method achieves lowest joint error relative to ground-truth execution (0.162~rad) and the lowest foot slip (10.8~cm/s) among the compared methods, reducing foot slip by 24.5\% relative to the best baseline.
These gains under the same fixed controller support our modeling of human motion and robot actions: executed-action supervision from closed-loop simulation captures control requirements of inter-frame transitions, while bidirectional kinematic alignment connects action prediction with motion reconstruction.

\paragraph{Performance across motion speeds.}
The advantage of robot mode is concentrated in faster motions (\cref{fig:robotmode}).
On slow motions, it performs similarly to human mode and slightly below GENMO.
On the fastest third by joint speed, however, its success reaches \intensJointOursRobotModeFast\%, compared with \intensJointGENMOFast\% for GENMO, and also exceeds human mode and the ground-truth references.
Across methods, the correlation analysis further indicates that human motion reconstruction errors alone do not reliably characterize execution success.
This pattern is consistent with our transition-aware formulation: executed-action targets supervise the controller commands required throughout each transition, complementing kinematic reconstruction when recovering motion for humanoid imitation.

\subsection{Ablation Study}
\label{sec:exp_ablation}

\paragraph{Protocol.}
\cref{fig:ablation_main} isolates the proposed components under executed-action supervision.
All variants share the data, human losses, optimizer, batch size, budget (50k steps), and seed of \method, changing only the listed components.
Evaluation follows \cref{sec:exp_human,sec:exp_robot}, with identical retargeting, root correction, controller, and \nSegAll\ segments.
We report robot-mode execution success, set-point MAE, camera-space MPJPE on 3DPW and EMDB-1, and world-space W-MPJPE$_{100}$ on EMDB-2.
MAE is measured on the training stream over steps 45k--50k.

\paragraph{Transition-aware modeling.}
Replacing the action branch's temporal blocks with a parameter-matched per-frame MLP raises set-point MAE from $5.49^\circ$ to $7.02^\circ$ and lowers robot-mode success from $86.0\%$ to $83.5\%$, even though the MLP still receives temporally contextualized backbone tokens.
With the temporal blocks retained, holding one set-point across all five sub-intervals of a transition ($K{=}1$) raises MAE to $5.86^\circ$ and lowers success to $83.1\%$.
These comparisons support using neighboring frames to infer commands a transition requires and predicting multiple set-points to capture how those commands vary within it.

\paragraph{Simple regression \vs proposed components.}
Starting from a per-frame action MLP on shared tokens, we successively add adapters, the temporal action branch, and alignment (\cref{fig:ablation_main}a).
Robot-mode success rises from $79.9\%$ to $86.0\%$, and set-point MAE falls from $7.42^\circ$ to $5.49^\circ$.
Adding the temporal branch gives the largest stepwise gain in success ($3.0$ percentage points) and reduction in MAE ($1.56^\circ$).
On 3DPW, MPJPE stays at $54.8$\,mm through the first two additions and falls to $53.7$\,mm with alignment.
The adapters and temporal decoder specialize shared features for command prediction, while alignment lets the robot branch's kinematic errors guide human pose reconstruction.

\paragraph{Bidirectional kinematic alignment.}
Removing $\cL_{\text{align}}$ lowers robot-mode success by $1.3$ percentage points and raises reconstruction error on all three benchmarks.
Both single-direction variants match the full model's $5.49^\circ$ set-point MAE, but robot-to-human and human-to-robot alignment yield success rates of $84.9\%$ and $83.7\%$, respectively, compared with $86.0\%$ using both directions.
Aligning absolute positions lowers success to $82.6\%$ and raises 3DPW MPJPE from $53.7$ to $55.6$\,mm.
Alternating teacher roles lets each branch guide the other with errors measured relative to its own labels, even when retargeting and controller corrections give the two branches different reference poses.

\paragraph{Action gradients in the shared backbone.}
To isolate the effect of action gradients, we compare the \emph{+ temporal branch} and \emph{detached} variants, both trained without alignment.
Allowing $\cL_{\text{act}}$ to update the backbone lowers set-point MAE from $5.91^\circ$ to $5.59^\circ$ and raises robot-mode success from $83.3\%$ to $84.7\%$.
The full model with bidirectional alignment achieves lower reconstruction errors than either variant on all three benchmarks.
The gradient path lets the set-points issued by the controller directly supervise the shared motion features, helping the backbone learn which commands will execute a transition successfully.

\subsection{Real-Robot Deployment}
\label{sec:exp_real}

We deploy \method on a physical LimX Luna humanoid~\cite{limx_luna} using same controller as simulation.
As shown in \cref{fig:real}, the robot completes both dance sequences recovered from in-the-wild videos without falling.
Using ViTPose keypoints, we measure robot's and dancer's limb angles relative to their respective torsos.
We compute frame-wise pose similarity by averaging cosines of differences between corresponding angles.
The sequence means are $0.77$ and $0.83$, respectively.
These results support our approach of coupling human motion recovery with executed-action supervision for real-world humanoid imitation.

\section{Conclusion}
\label{sec:conclusion}

We present \method, which jointly learns human motion recovery and robot action prediction in a single training stage.
Executed controller actions provide dense supervision for the control requirements of inter-frame transitions, while shared temporal modeling and bidirectional kinematic alignment enable mutual guidance between the two tasks.
A single forward pass supports human motion recovery and humanoid imitation without retarget.
Experiments demonstrate strong reconstruction accuracy and the highest overall execution success among the compared methods without test-time joint smoothing, with the largest gains on fast motions.
These results support learning reconstruction fidelity and execution requirements together through complementary human motion and robot action representations.

\paragraph{Limitations.}
Our executed-action supervision is specific to one robot and controller, and robot execution depends on the accuracy of the recovered human root trajectory.
Future work will investigate transfer across robots and controllers, explicit modeling of environmental contacts, and broader validation on physical humanoids.

\section*{LLM Usage Statement}
LLMs (ChatGPT and Claude) are used for language polishing and grammar correction in parts of the manuscript.
All scientific content, methodology, experiments, and conclusions are developed and verified by the authors.
{
    \small
    \bibliographystyle{ieeenat_fullname}
    \bibliography{main}
}


\end{document}